\documentclass[letterpaper]{article}

\usepackage{natbib,alifeconf}  
\usepackage{hyperref}

\title{Neural Network Quine}
\author{Oscar Chang$^{1}$ \and Hod Lipson$^{1}$ \\
\mbox{}\\
$^1$Data Science Institute, Columbia University, New York, NY 10027\\
{oscar.chang, hod.lipson}@columbia.edu} 

\usepackage{lipsum}
\usepackage{program}

\begin{document}
\maketitle
\begin{abstract}
  Self-replication is a key aspect of biological life that has been largely overlooked in Artificial Intelligence systems. Here we describe how to build and train self-replicating neural networks. The network replicates itself by learning to output its own weights. The network is designed using a loss function that can be optimized with either gradient-based or non-gradient-based methods. We also describe a method we call {\em regeneration} to train the network without explicit optimization, by injecting the network with predictions of its own parameters. The best solution for a self-replicating network was found by alternating between regeneration and optimization steps. Finally, we describe a design for a self-replicating neural network that can solve an auxiliary task such as MNIST image classification. We observe that there is a trade-off between the network's ability to classify images and its ability to replicate, but training is biased towards increasing its specialization at image classification at the expense of replication. This is analogous to the trade-off between reproduction and other tasks observed in nature. We suggest that a self-replication mechanism for artificial intelligence is useful because it introduces the possibility of continual improvement through natural selection.

\end{abstract}

\section{Introduction}
The concept of an artificial self-replicating machine was first proposed by John von Neumann in the 1940s prior to the discovery of DNA's role as the physical mechanism for biological replication. Specifically, Von Neumann demonstrated a configuration of initial states and transformation rules for a cellular automaton that produces copies of the initial cell states after running for a fixed number of steps \citep{neumann}. \citet{douglas} later coined the term `quine' in \textit{G{\"o}del, Escher, Bach: an Eternal Golden Braid} after the philosopher Willard Van Orman Quine, to describe self-replicating expressions such as: `is a sentence fragment' is a sentence fragment. 

There has also been work done in making physical self-replicators. Notable examples include tiles \citep{penrose}, molecules \citep{wang2011self}, polymers \citep{breivik2001self}, and robots \citep{lipson}.

In the context of programming language theory, quines are computer programs that print their own source code. A trivial example of a quine is the empty string, which in most languages, the compiler transforms into the empty string. The following code snippet is an example of a non-trivial Python quine written in two lines\nocite{wiki:quine}.

\begin{verbatim}
s = 's = %r\nprint(s%%s)'
print(s%s)
\end{verbatim}

While self-replication has been studied in many automata, it is notably absent in neural network research, despite the fact that neural networks appear to be the most powerful form of AI known to date. 

In this paper, we identify and attempt to solve the challenges involved in building and training a self-replicating neural network. Specifically, we propose to view a neural network as a differentiable computer program composed of a sequence of tensor operations. Our objective then is to construct a neural network quine that outputs its own weights.

We tested our approach using three distinct classes of methods: gradient-based optimization methods, non-gradient-based optimization methods, and a novel method called regeneration. We further designed a neural network quine which has an auxiliary objective in addition to the job of self-replication. In this paper, the chosen auxiliary task is MNIST image classification \citep{mnist}, which involves classifying images of digits from 0 to 9, and is commonly used as a `hello world' example for machine learning.

We observed a trade-off between the network's ability to self-replicate and its ability to solve the auxiliary task. This is analogous to the trade-off between reproduction and other auxiliary survival tasks observed in nature. The two objectives are usually aligned, but for example, when an animal has been put in starving conditions, its sex hormones are usually down-regulated to optimize for survival at the expense of reproduction. The opposite occurs as well: for example, in male dark fishing spiders, the act of copulation results in a sudden irreversible change to its blood pressure, immobilizing it and leaving it vulnerable to cannibalization by the female spider \citep{spider}.

\subsection{Motivations}
Modern artificial intelligence is primarily powered by deep neural networks for applications as diverse as detecting diabetic retinopathy \citep{diabetes}, synthesizing human-like speech \citep{shen2017natural}, and executing strategic decisions in Starcraft \citep{vinyals2017starcraft}. In line with ALIFE 2018's theme of going \textit{beyond AI}, we list several motivations for studying self-replicating neural networks.
\begin{itemize}
  \item Biological life began with the first self-replicator \citep{newscientist}, and natural selection kicked in to favor organisms that are better at replication, resulting in a self-improving mechanism. Analogously, we can construct a self-improving mechanism for artificial intelligence via natural selection if AI agents had the ability to replicate and improve themselves without additional machinery.
  \item Neural networks are capable of learning powerful representations across many different domains of data \citep{representation}. But can a neural network learn a good representation of itself? Self-replication involves a degree of introspection and self-awareness, which is helpful for lifelong learning and potential discovery of new neural network architectures.
  \item Learning how to enhance or diminish the ability for AI programs to self-replicate is useful for computer security. For example, we might want an AI to be able to execute its source code without being able to read or reverse-engineer it, either through its own volition or interaction with an adversary.
  \item Self-replication functions as the ultimate mechanism for self-repair in damaged physical systems \citep{lipson}. The same may apply to AI, where a self-replication mechanism can serve as the last resort for detecting damage, or returning a damaged or out-of-control AI system back to normal.
\end{itemize}

\subsection{Related Work}
Quines have been written for a variety of programming languages. The Quine Page \citep{quine_thompson} contains code contributions of quines written in $55$ different languages. An Ouroboros set of programs extends the concept of a quine by having a program in language A generate the source code for a program in language B, which then generates the source code for a program in a language C, and so on, until it finally generates back the source code for the initial program in language A. \citet{quine_relay} made an Ouroboros with $128$ programming languages in it.

Our work focuses on building a self-replication mechanism via weight prediction. \citet{denil2013predicting} demonstrated the presence of redundancy in neural networks by using a portion of the weights to predict the rest. There are also neural networks that can modify the weights of other neural networks \citep{schmidhuber1992learning,hypernetwork}, which have been shown to be useful in meta-learning an optimizer \citep{ravi2016optimization, andrychowicz2016learning}. \citet{schmidhuber1993self} proposed an architecture and a training algorithm for a self-referential recurrent neural network, which is philosophically very similar to our work in that the network refers to itself rather than another network. To our knowledge, our work is the first to attempt the task of self-replication in neural networks.

\section{Building the Network}
\subsection{How can a neural network refer to itself?}
\subsubsection{Problem with Direct Reference}
A neural network is parametrized by a set of parameters $\Theta$, and our goal is to build a network that outputs $\Theta$ itself. This is difficult to do directly. Suppose the last layer of a feed-forward network has $A$ inputs and $B$ outputs. Already, the size of the weight matrix in a linear transformation is the product $AB$ which is greater than $B$ for any $A>1$.

We also looked at open-source implementations of two popular generative models for images, DCGAN \citep{dcgan} and DRAW \citep{draw}. They use 12 million and 1 million parameters respectively to generate MNIST images with 784 pixels.

In general, the set of parameters $\Theta$ is a lot larger than the size of the output. To circumvent this, we need an indirect way of referring to $\Theta$.

\subsubsection{Indirect Reference}
HyperNEAT \citep{hyperneat} is a neuro-evolution method that describes a neural network by identifying every topological connection with a coordinate and a weight. We pursue the same strategy in building a quine. Instead of having the quine output its weights directly, we shall set it up so that it inputs a coordinate (in a one-hot encoding) and outputs the weight at that coordinate.

This overcomes the problem of $\Theta$ being larger than the output, since we are only outputting a scalar $\Theta_c$ for each coordinate $c$.

\subsection{Vanilla Quine}
We define the \textit{vanilla quine} as a feed-forward neural network whose only job is to output its own weights.

\begin{figure}[h]
\begin{center}
\includegraphics[width=0.5\textwidth]{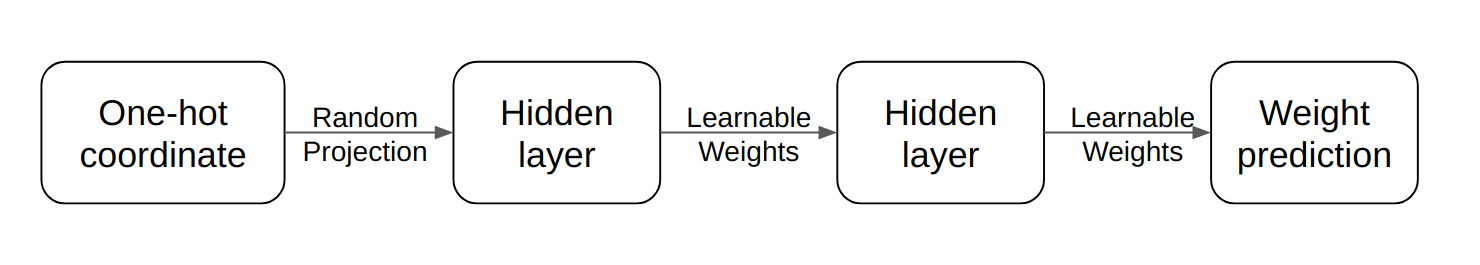}
\end{center}
\caption{Structure of a vanilla quine}
\label{fig:vquine_diagram}
\end{figure}

Suppose the number of inputs is $A$, and the number of units in the first hidden layer is $B$, then the size of the projection matrix would be the product $AB$ which is greater than $A$ for any $B>1$. Hence, we cannot have the projection itself be a parameter of the network due to the one-hot representation. We thus decide to use a fixed random projection to connect the one-hot encoding of the coordinate to the hidden layer. All other connections, namely the connections between the hidden layers as well as the connections between the last hidden layer and the output layer, are variable parameters of the neural network.

Von Neumann \nocite{wiki:neumann}argued that a non-trivial self-replicator necessarily includes three components that by themselves do not suffice to be self-replicators:\ (1) a description of the replicator, (2) a copying mechanism that can clone descriptions, and (3) a mechanism that can embed the copying mechanism within the replicator itself \citep{neumann}. In this case, the coordinate system that assigns each of the weights a point in the one-hot space corresponds to (1). The function computed by the neural network corresponds to (2). The fixed random projection corresponds to (3). We explain below reasons for our choices of (1), (2), and (3), while keeping in mind that alternatives to these choices are interesting future research directions.

\subsubsection{(1) One-hot Input Encoding}

\begin{figure}[h]
\begin{center}
\includegraphics[width=0.25\textwidth]{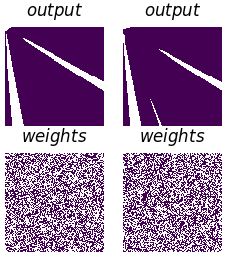}
\end{center}
\caption{Log-normalized illustration of the weights and weight predictions of a quine without one-hot encoding}
\label{fig:no-one-hot}
\end{figure}

A one-hot encoding is a vector that contains exactly a single '1', and is 0 everywhere else. If we directly input the coordinate instead of using a one-hot encoding, then the network will not be sufficiently expressive. This is because for any coordinate $c$, the difference between $f(c)$ and $f(c+1)$ is constrained by the network's Lipschitz bound, hence the network cannot accurately output the weights at $c$ and $c+1$ if their difference is sufficiently big. We demonstrate a visualization of this in Figure \ref{fig:no-one-hot}: contiguous weights might be very different, but contiguous outputs cannot be very different.

\subsubsection{(2) Multi-Layered Perceptrons}
\begin{equation}\label{eq:1}
    y_i = \sigma_i (W_ix_i+b_i)
    \label{eqn:mlp}
\end{equation}

Multi-layered perceptrons (MLPs) are feed-forward neural networks that consist of repeated applications of Equation \ref{eqn:mlp}, where at the $i$th layer of the network, $\sigma_i$ is an activation function, $W_i$ a weight matrix, $b_i$ a bias vector, $x_i$ the input vector, and $y_i$ the output vector. MLPs are known to be good function approximators, specifically a feedforward neural network with at least one hidden layer forms a class of functions that is dense in the space of continuous functions under a compact domain \citep{hornik1991approximation, cybenko1989approximation}. While not precluding other kinds of generative neural network architectures, this makes an MLP seem like a suitable candidate for a neural network quine, because we think it is expressive enough to derive and store a representation of itself.

\subsubsection{(3) Random Projections}
We believe that random projections are a good choice as an embedding layer to connect a one-hot representation into the network because of their distance-preserving property \citep{johnson1984extensions} and the fact that random features have been shown to work well both in theory and practice \citep{rahimi2008random}. Indeed, they form a key component of Extreme Learning Machines \citep{huang2006extreme} which are feed-forward neural networks that have proven useful in classification and regression problems.

\subsection{Auxiliary Quine}
We define the \textit{auxiliary quine} to be a vanilla quine that solves some auxiliary task in addition to self-replication. It is responsible for taking in an auxiliary input and returning an auxiliary output.

\begin{figure}[h]
\begin{center}
\includegraphics[width=0.5\textwidth]{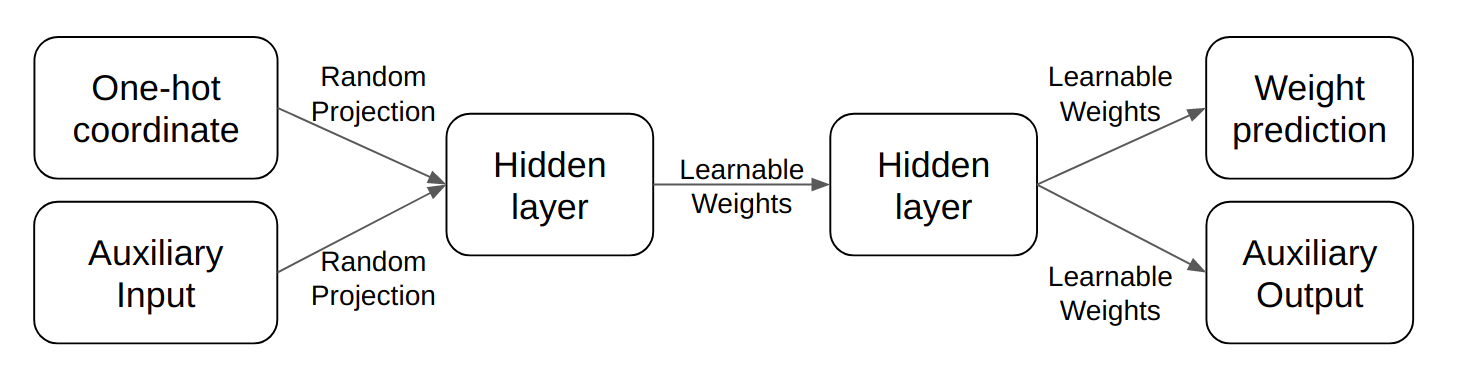}
\end{center}
\caption{Structure of an auxiliary quine}
\label{fig:mquine_diagram}
\end{figure}

In this paper, we chose image classification as the auxiliary task. The MNIST dataset \citep{mnist} contains square images (28 pixels by 28 pixels) of handwritten digits from 0 to 9. These images are to be fed in as the auxiliary input. It is possible to make the connection from the auxiliary input to the network a parameter rather than a random projection, but in this paper, we only report results for the latter. 

The auxiliary output is a probability distribution over the ten classes, where the class with the maximum probability will be chosen as the predicted classification. $60000$ images are used for training and $10000$ images are used for testing; we have no need for a validation set since we are not strictly trying to optimize for the performance of the classifier. Our primary aim in this paper is to demonstrate a proof of concept for a neural network quine, which makes MNIST a suitable auxiliary task as it is considered an easy problem for modern machine learning algorithms.

\section{Training the Network}
\subsection{Network Architecture}
Before describing how the neural network quines are trained, we specify the exact network architecture used in our experiments below for both the vanilla quine and the auxiliary quine. In both cases, they are MLPs composed of two hidden layers with 100 hidden units each where every layer is followed by a SeLU \citep{klambauer2017self} activation function. In the case of the auxiliary quine, the one-hot coordinate is projected to the first 50 hidden units, while the MNIST input is projected to the next 50 hidden units. The auxiliary output is a vector of size $10$ (number of classes) computed by a softmax.

The total number of parameters is 20,100 for the vanilla quine, and 21,100 for the auxiliary quine. The nature of the quine problem and our choice of the one-hot encoding means that the input vector will be of the same size as the number of parameters. These are small networks by modern deep learning standards where millions of parameters are the norm, but it is a challenge to handle input vectors with dimensions much larger than 20,000.

\subsection{How do we train a neural network quine?}
\subsubsection{Self-Replicating Loss}
We define the \textit{self-replicating loss} to be the sum of the squared difference between the actual weight and its predicted value. A vanilla quine is achieved when this loss is exactly zero. Because of numerical imprecision errors, we can expect that in practice, optimizing this loss will nonetheless result in a number slightly above zero, except for the trivial \textit{zero quine} where all the weights are exactly zero to begin with.
\begin{equation}
    L_{SR} = \sum_{c \in C} \Big\|f_\Theta(c)-\Theta_c\Big\|^2_2
\label{eqn:sr-loss}
\end{equation}
\subsubsection{Auxiliary Loss}
It  is  possible  to  jointly  optimize  an  existing  loss  function with the self-replicating loss so that a neural network gains the ability to self-replicate in addition to an auxiliary task it specializes in. We define the \textit{auxiliary loss} to be the sum of the self-replicating loss $L_{SR}$ and the loss from the auxiliary task $L_{Task}$, with a hyperparameter $\lambda$ to scale both losses to a similar magnitude. An auxiliary quine can be trained by optimizing on the auxiliary loss, but we do not expect to see a near-zero loss, unless it is also perfect at the auxiliary task. In our MNIST experiment, $L_{Task}$ is the cross-entropy loss, which is commonly used for classification problems.
\begin{equation}
    L_{Aux} = L_{SR} + \lambda L_{Task}
\label{eqn:aux-loss}
\end{equation}
\subsubsection{Training Methods}
There are three distinct classes of methods that we can use to train our neural network quines.
\begin{itemize}
    \item \textbf{Gradient-based methods} Stochastic gradient descent (SGD) and its variants are the workhorse algorithm for training deep neural networks today. In our case, the loss function is a moving target, since $\Theta_c$ changes after each gradient update. Updating the loss function after every mini-batch update is expensive. To avoid that, we split the set of possible coordinates into random mini-batches of size 10, and update the loss function after every training epoch. In other words, each training epoch will consist of running through the set of all possible coordinates. We do not use a validation set for our experiments, while the test loss is computed at the end of every training epoch after updating the loss function. Below is pseudo-code for training a vanilla quine. A similar procedure is used to train an auxiliary quine with $L_{SR}$ replaced with $L_{Aux}$ to account for the auxiliary task.
    \begin{program}
    \mbox{Pseudo-code for training a vanilla quine via optimization:}
    \BEGIN \\ %
      |Initialize set of parameters|\ \Theta_C
      |Initialize number of training epochs|\ T
      \FOR t:=0 \TO T \DO
        \Theta_t := \Theta_C
        |Divide|\ \Theta_t\ |into random mini-batches|
        \FOR |each mini-batch| \DO
            |Compute|\ L_{SR}
            \Theta_C := |optimize(|\Theta_C|,|\ L_{SR}|)|
        \END
      \END
    \END
    \label{alg:1}
    \end{program}
    \item \textbf{Non-gradient-based methods} Optimization methods that do not make use of gradient information can also be used to train neural networks. For example, evolutionary algorithms have been used successfully to train deep reinforcement learning agents with over four million parameters \citep{such2017deep, salimans2017evolution}. For the same reasons of computational efficiency as mentioned above, we shall choose to execute non-gradient-based optimization in mini-batches. The training algorithm is identical to the pseudo-code shown above, except with \texttt{optimize} being non-gradient-based. We only consider hill-climbing in this paper.
    \item \textbf{Regeneration method} Perhaps somewhat surprisingly, it is also possible to train a vanilla quine without explicitly optimizing for the self-replicating loss. We do so by replacing the current set of parameters with the weight predictions made by the quine. Each such replacement is called a \textit{regeneration}. We then alternate between executing regeneration and a round of optimization to achieve a low but non-trivial self-replicating loss. We note that regeneration is sensitive to choices of weight initialization and activation function.
    \begin{program}
    \mbox{Pseudo-code for Regeneration:}
    \BEGIN \\ %
      |Initialize set of parameters|\ \Theta_C
      |Initialize number of generation epochs|\ G
      |Initialize number of optimization epochs|\ T
      \FOR g:=0 \TO G \DO
          //\ Optimization
          \FOR t:=0 \TO T \DO
            \Theta_t := \Theta_C
            |Divide|\ \Theta_t\ |into random mini-batches|
            \FOR |each mini-batch| \DO
                |Compute|\ L_{SR}
                \Theta_C := |optimize(|\Theta_C|,|\ L_{SR}|)|
            \END
          \END
          //\ Regeneration
          \FOR c\in C \DO
            \Theta_c := f_{\Theta_C}(c)
          \END
      \END
    \END
    \label{alg:1}
    \end{program}
\end{itemize}

\section{Results and Discussion}
In the experimental results produced below, we used a mini-batch of size $10$ for training. $\lambda$ in $L_{Aux}$ and the temperature for the softmax in the auxiliary output are set to $0.01$.

\subsection{Vanilla Quine}
\begin{figure}[h]
\begin{center}
\includegraphics[width=0.5\textwidth]{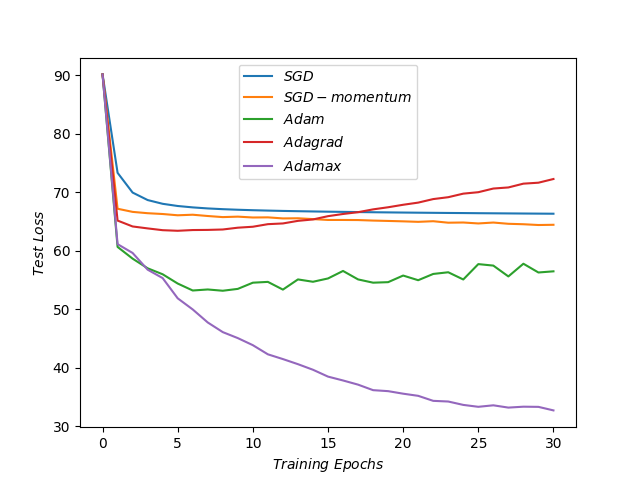}
\end{center}
\caption{Comparison of gradient-based optimization methods used to train a vanilla quine}
\label{fig:vquine_gradient}
\end{figure}

We trained a vanilla quine with classical SGD ($lr=0.01$), SGD with momentum ($lr=0.01,\ \rho=0.9$), ADAM \citep{kingma2014adam}, Adagrad \citep{duchi2011adaptive}, Adamax \citep{kingma2014adam}, and RMSprop \citep{rmsprop} with default hyperparameter settings on the self-replicating loss for $30$ epochs. The quine was initialized with the same procedure as in \citet{he2015delving}, and the initial loss $L_{SR}$ prior to any training was $90.16$. We observe in Figure \ref{fig:vquine_gradient} that Adamax performed the best, while Adagrad exhibited increasing loss rather than plateauing. RMSprop (not plotted) was found to explode the loss right from the start of training. We carried on training the quine on Adamax for $100$ epochs, achieving a best test loss of $32.10$ by the end of training, which is a third of its pre-trained value.

\subsection{Is this a quine?}

It is hard to quantify how significant it is to reduce the self-replicating loss to a third of its pre-trained value. After all, our goal was to produce a self-replicator, but if the loss we achieved is not close to zero, then it seems that we have not reached our goal. On the other hand, replication mechanisms are rarely perfect. Even in nature, replication mechanisms often contain high levels of noise, sometimes referred to as `mutation' or replication error.

\citet{quotient} constructed a mathematical framework to calculate the \textit{self-replicating quotient} of a replicator, which measures the log likelihood ratio of a perfect self-replication happening via the replicator's noisy replication mechanism compared to it happening by chance. For example, \cite{lipson} estimate the self-replicating quotient of Penrose Tiling \citep{penrose} to be below $0.69$ and that of animals to be at least $46.05$. This framework is useful for distinguishing between trivial and non-trivial replicators. To compute this metric for our network, we need to compute the chance that a random neural network would produce a copy of itself within the same $\epsilon$-ball as achieved by our vanilla quine. Assuming that a random network has a uniform distribution of outputs from $\qquad[-0.5, 0.5]$ (a big assumption), then the self-replicating quotient for the vanilla quine is $6.44$, which implies a certain amount of non-triviality.

Another measure we can look at is the \textit{average weight prediction margin}, which is defined as the average absolute difference between the weights and the weight predictions. The pre-training loss of $90.16$ corresponds to an average weight prediction margin of 0.067, while the post-training loss of $32.10$ corresponds to an average weight prediction margin of 0.040. This suggests we still have significant room for improvement. However, it is worth pointing out that the relatively small pre-training weight prediction margin reflects the fact that modern best practices for the choice of weight initialization and activation function keep the output in the same order of magnitude as the input.

\subsection{Hill-climbing}

Next, we use a hill-climbing algorithm to train the vanilla quine. The algorithm works by iteratively perturbing the parameters of the network with diagonal Gaussian noise and keeping the perturbation if it results in an improvement. This is equivalent to an evolutionary algorithm with a population size of $1$. In this case, we do not need the gradients, hence the training process only requires the forward and not the backward pass, which makes each training epoch computationally cheaper. Nonetheless, it takes around $5000$ epochs to find a solution that is on par with that found by classical SGD after $10$ epochs. We found that doing hill-climbing on the solution that SGD converged to improves it significantly, but the same does not hold true for the solution that Adamax converged to. This suggests that the solution found by Adamax is already a local optima.

\begin{figure}[h]
\begin{center}
\includegraphics[width=0.5\textwidth]{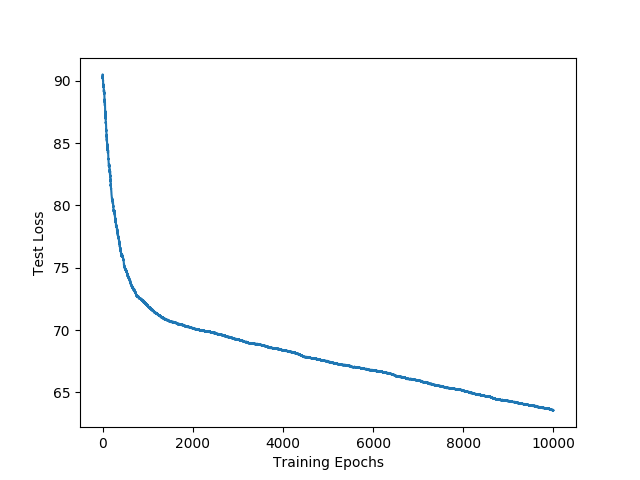}
\end{center}
\caption{Training a vanilla quine via hill-climbing}
\label{fig:vquine_evo}
\end{figure}

\subsection{Regeneration}

Finally, we use regeneration to train the vanilla quine, setting $T=1$ with Adamax as the optimizer. Each generation epoch is very computationally expensive as it involves as many forward passes as there are parameters in the network to replace its actual weights with its outputs. However, one epoch suffices to reduce the test loss substantially, with the best self-replicating loss of $0.86$ found after ten generation epochs. This corresponds to a self-replicating quotient of $10.06$ and an average weight prediction margin of $0.0065$, which is an order of magnitude better than the best solution found previously.

\begin{figure}[h]
\begin{center}
\includegraphics[width=0.5\textwidth]{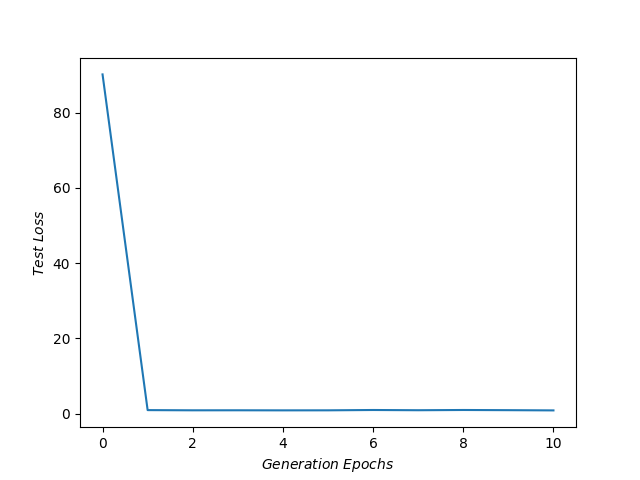}
\end{center}
\caption{Training a vanilla quine via regeneration}
\label{fig:vquine_gen}
\end{figure}

One might wonder if the solution we found via regeneration might simply be the zero quine, i.e.\ all the weights are zero. Indeed, we find that iteratively injecting the network with its predicted weights has a similar effect as statistical shrinkage. It effectively learns to reduce the self-replicating loss by shrinking the order of magnitudes of the weights, thus creating a small weight prediction margin. Without the optimization step (when $T=0$), a visual inspection of the network reveals that it rapidly converges to the zero quine. However, with the optimization step, the solution found appears to be non-trivial: the order of magnitude of the weights are in line with what we would observe in a normal neural network. The strong attraction of the zero quine also underscores the importance of having an additional auxiliary task.

\begin{figure}[h]
\begin{center}
\includegraphics[width=0.3\textwidth]{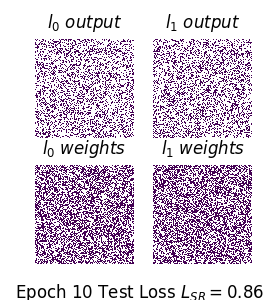}
\end{center}
\caption{Log-normalized illustration of the weights and weight predictions of two hidden layers in a vanilla quine that has been trained with regeneration}
\label{fig:vquine_res}
\end{figure}

Figure \ref{fig:vquine_res} shows a visualization of the solution found by regeneration.

\subsection{Auxiliary Quine}

\begin{figure}[h]
\begin{center}
\includegraphics[width=0.5\textwidth]{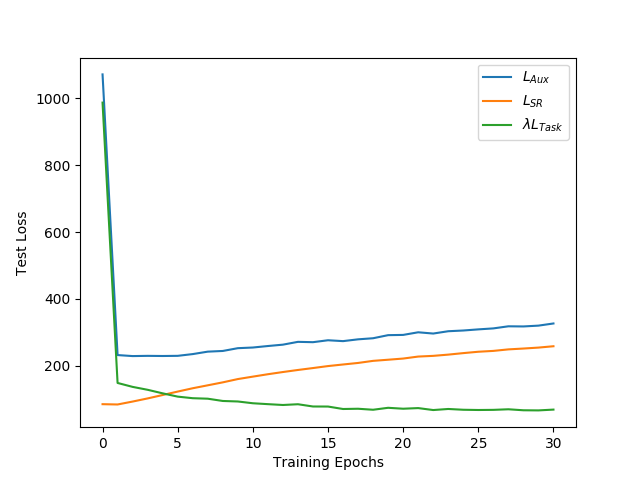}
\end{center}
\caption{Training an auxiliary quine with Adamax}
\label{fig:mquine_loss}
\end{figure}

We trained an auxiliary quine on the MNIST image classification task with Adamax using the default hyperparameter settings on $30$ epochs. The quine was also initialized with He init, and the initial loss $L_{Aux}$ prior to any training was $1072.05$. We observe in Figure \ref{fig:mquine_loss} that somewhat counter-intuitively, after the initial drop, the auxiliary loss actually increases over time instead of converging. This is due to the network prioritizing the task loss $L_{Task}$ over the self-replicating loss $L_{SR}$ despite the fact that it is being optimized on their sum. The same trend is observed when we repeat the experiment on other gradient-based optimization methods besides Adamax. After $30$ epochs, the network achieved an accuracy of $90.41\%$ on the held-out test set, which is comparable to the $96.33\%$ achieved by an identical network whose only objective is MNIST image classification. This shows that self-replication occupies a significant portion of the neural network's capacity, but it is heartening nonetheless that joint optimization of the objectives is possible. If we leave the auxiliary quine running, the task loss eventually converges, while ignoring the exploding self-replicating loss.

This is an interesting finding: it is more difficult for a network that has increased its specialization at a particular task to self-replicate. This suggests that the two objectives are at odds with each other, but that the gradient-based optimization procedure prefers to maximize the network's specialization at solving the MNIST task, even at the expense of a reduction in its ability to self-replicate. (It is not immediately obvious from Figure \ref{fig:mquine_loss}, but the first few training epochs reduce the self-replicating loss too.)

There are parallels to be drawn between self-replication in the case of a neural network quine and biological reproduction in nature, as well as specialization at the auxiliary task and survival in nature. The mechanisms for survival are usually aligned with the mechanisms for reproduction, however when they come into conflict with each other, the survival mechanism usually is prioritized at the expense of the reproduction mechanism (except in rare cases like that of the male dark fishing spider). We hypothesize that in nature, self-replication might rapidly become trivial without the presence of an auxiliary task, just as we have observed here in the case of a neural network quine.

Hill-climbing progressed too slowly for us to observe anything meaningful, but we do not expect to observe the same behavior because the algorithm, by definition, does not allow for harmful changes to the overall loss to be made. The regeneration techique cannot be used in this case, because we require the auxiliary input for each generation and random inputs do not work well.

\section{Conclusion}

In this paper, we have described how to build and train a self-replicating neural network. Specifically, we proposed to treat the problem of self-replication in a neural network as a problem of weight prediction, and devised various encoding and training schemes to solve this problem. This allowed us to create a neural network quine, which akin to a computer program quine, outputs its own source code (weights in this case).

We have identified three interesting future directions for research. Firstly, we can seek to improve weight prediction by assuming a low-rank matrix factorization for the network's weights as in \citet{denil2013predicting}. Secondly, we can attempt to build neural network quines using more sophisticated models and representations, for example a convolutional neural network quine might be interesting. Thirdly, we can extend the concept of self-replication to universal replication: a neural network that can replicate other neural networks.

\section{Acknowledgements}

We would like to thank Peter Duraliev for helpful comments. This research was supported in part by the US Defense Advanced Research Project Agency (DARPA) \textit{Lifelong Learning Machines} Program, grant HR0011-18-2-0020.

\footnotesize
\bibliographystyle{apalike}
\bibliography{example} 

\begin{thebibliography}{}

\bibitem[Adams and Lipson, 2009]{quotient}
Adams, B. and Lipson, H. (2009).
\newblock A universal framework for analysis of self-replication phenomena.
\newblock {\em Entropy}, 11:295--325.

\bibitem[Andrychowicz et~al., 2016]{andrychowicz2016learning}
Andrychowicz, M., Denil, M., Gomez, S., Hoffman, M.~W., Pfau, D., Schaul, T.,
  and de~Freitas, N. (2016).
\newblock Learning to learn by gradient descent by gradient descent.
\newblock In {\em Advances in Neural Information Processing Systems}, pages
  3981--3989.

\bibitem[Bengio et~al., 2013]{representation}
Bengio, Y., Courville, A., and Vincent, P. (2013).
\newblock Representation learning: A review and new perspectives.
\newblock {\em IEEE transactions on pattern analysis and machine intelligence},
  35, no. 8:1798--1828.

\bibitem[Breivik, 2001]{breivik2001self}
Breivik, J. (2001).
\newblock Self-organization of template-replicating polymers and the
  spontaneous rise of genetic information.
\newblock {\em Entropy}, 3(4):273--279.

\bibitem[contributors, 2017]{wiki:neumann}
contributors, W. (2017).
\newblock Self-replication --- wikipedia{,} the free encyclopedia.
\newblock [Online; accessed 5-March-2018].

\bibitem[contributors, 2018]{wiki:quine}
contributors, W. (2018).
\newblock Quine (computing) --- wikipedia{,} the free encyclopedia.
\newblock [Online; accessed 5-February-2018].

\bibitem[Cybenko, 1989]{cybenko1989approximation}
Cybenko, G. (1989).
\newblock Approximation by superpositions of a sigmoidal function.
\newblock {\em Mathematics of control, signals and systems}, 2(4):303--314.

\bibitem[Denil et~al., 2013]{denil2013predicting}
Denil, M., Shakibi, B., Dinh, L., De~Freitas, N., et~al. (2013).
\newblock Predicting parameters in deep learning.
\newblock In {\em Advances in neural information processing systems}, pages
  2148--2156.

\bibitem[Drake, 2013]{spider}
Drake, N. (2013).
\newblock Why male dark fishing spiders die spontaneously after sex.
\newblock {\em Wired}.

\bibitem[Duchi et~al., 2011]{duchi2011adaptive}
Duchi, J., Hazan, E., and Singer, Y. (2011).
\newblock Adaptive subgradient methods for online learning and stochastic
  optimization.
\newblock {\em Journal of Machine Learning Research}, 12(Jul):2121--2159.

\bibitem[Endoh, 2017]{quine_relay}
Endoh, Y. (2017).
\newblock Quine relay.

\bibitem[Gregor et~al., 2015]{draw}
Gregor, K., Danihelka, I., Graves, A., Rezende, D.~J., and Wierstra, D. (2015).
\newblock Draw: A recurrent neural network for image generation.
\newblock {\em Proceedings of the 32nd International Conference on Machine
  Learning}.

\bibitem[Gulshan et~al., 2016]{diabetes}
Gulshan, V., Peng, L., Coram, M., Stumpe, M.~C., Wu, D., Narayanaswamy, A.,
  Venugopalan, S., Widner, K., Madams, T., Cuadros, J., Kim, R., Raman, R.,
  Nelson, P.~C., Mega, J.~L., and Webster, D.~R. (2016).
\newblock Development and validation of a deep learning algorithm for detection
  of diabetic retinopathy in retinal fundus photographs.
\newblock {\em JAMA}, 316(22):2402--2410.

\bibitem[Ha et~al., 2017]{hypernetwork}
Ha, D., Dai, A.~M., and Le, Q.~V. (2017).
\newblock Hypernetworks.
\newblock {\em International Conference on Learning Representations}.

\bibitem[He et~al., 2015]{he2015delving}
He, K., Zhang, X., Ren, S., and Sun, J. (2015).
\newblock Delving deep into rectifiers: Surpassing human-level performance on
  imagenet classification.
\newblock In {\em Proceedings of the IEEE international conference on computer
  vision}, pages 1026--1034.

\bibitem[Hofstadter, 1980]{douglas}
Hofstadter, D. (1980).
\newblock {\em G{\"o}del, Escher, Bach: an Eternal Golden Braid}.
\newblock New York: Vintage Books.

\bibitem[Hornik, 1991]{hornik1991approximation}
Hornik, K. (1991).
\newblock Approximation capabilities of multilayer feedforward networks.
\newblock {\em Neural networks}, 4(2):251--257.

\bibitem[Huang et~al., 2006]{huang2006extreme}
Huang, G.-B., Zhu, Q.-Y., and Siew, C.-K. (2006).
\newblock Extreme learning machine: theory and applications.
\newblock {\em Neurocomputing}, 70(1-3):489--501.

\bibitem[Johnson and Lindenstrauss, 1984]{johnson1984extensions}
Johnson, W.~B. and Lindenstrauss, J. (1984).
\newblock Extensions of lipschitz mappings into a hilbert space.
\newblock {\em Contemporary mathematics}, 26(189-206):1.

\bibitem[Kingma and Ba, 2014]{kingma2014adam}
Kingma, D.~P. and Ba, J. (2014).
\newblock Adam: A method for stochastic optimization.
\newblock {\em arXiv preprint arXiv:1412.6980}.

\bibitem[Klambauer et~al., 2017]{klambauer2017self}
Klambauer, G., Unterthiner, T., Mayr, A., and Hochreiter, S. (2017).
\newblock Self-normalizing neural networks.
\newblock In {\em Advances in Neural Information Processing Systems}, pages
  972--981.

\bibitem[LeCun and Cortes, 1998]{mnist}
LeCun, Y. and Cortes, C. (1998).
\newblock The mnist database of handwritten digits.

\bibitem[Marshall, 2011]{newscientist}
Marshall, M. (2011).
\newblock First life: The search for the first replicator.
\newblock {\em New Scientist}, Issue 2825.

\bibitem[Penrose, 1959]{penrose}
Penrose, L.~S. (1959).
\newblock Self-reproducing machines.
\newblock {\em Scientific American}, 200:105--112.

\bibitem[Radford et~al., 2016]{dcgan}
Radford, A., Metz, L., and Chintala, S. (2016).
\newblock Unsupervised representation learning with deep convolutional
  generative adversarial networks.
\newblock {\em International Conference on Learning Representations}.

\bibitem[Rahimi and Recht, 2008]{rahimi2008random}
Rahimi, A. and Recht, B. (2008).
\newblock Random features for large-scale kernel machines.
\newblock In {\em Advances in neural information processing systems}, pages
  1177--1184.

\bibitem[Ravi and Larochelle, 2016]{ravi2016optimization}
Ravi, S. and Larochelle, H. (2016).
\newblock Optimization as a model for few-shot learning.

\bibitem[Salimans et~al., 2017]{salimans2017evolution}
Salimans, T., Ho, J., Chen, X., and Sutskever, I. (2017).
\newblock Evolution strategies as a scalable alternative to reinforcement
  learning.
\newblock {\em arXiv preprint arXiv:1703.03864}.

\bibitem[Schmidhuber, 1992]{schmidhuber1992learning}
Schmidhuber, J. (1992).
\newblock Learning to control fast-weight memories: An alternative to dynamic
  recurrent networks.
\newblock {\em Neural Computation}, 4(1):131--139.

\bibitem[Schmidhuber, 1993]{schmidhuber1993self}
Schmidhuber, J. (1993).
\newblock A ‘self-referential’weight matrix.
\newblock In {\em ICANN’93}, pages 446--450. Springer.

\bibitem[Shen et~al., 2017]{shen2017natural}
Shen, J., Pang, R., Weiss, R.~J., Schuster, M., Jaitly, N., Yang, Z., Chen, Z.,
  Zhang, Y., Wang, Y., Skerry-Ryan, R., et~al. (2017).
\newblock Natural tts synthesis by conditioning wavenet on mel spectrogram
  predictions.
\newblock {\em arXiv preprint arXiv:1712.05884}.

\bibitem[Stanley et~al., 2009]{hyperneat}
Stanley, K.~O., D’Ambrosio, D., and Gauci, J. (2009).
\newblock A hypercube-based indirect encoding for evolving large-scale neural
  networks.
\newblock {\em Artificial Life}, 15(2):185--212.

\bibitem[Such et~al., 2017]{such2017deep}
Such, F.~P., Madhavan, V., Conti, E., Lehman, J., Stanley, K.~O., and Clune, J.
  (2017).
\newblock Deep neuroevolution: Genetic algorithms are a competitive alternative
  for training deep neural networks for reinforcement learning.
\newblock {\em arXiv preprint arXiv:1712.06567}.

\bibitem[Thompson, 1999]{quine_thompson}
Thompson, G.~P. (1999).
\newblock The quine page.

\bibitem[Tieleman and Hinton, 2012]{rmsprop}
Tieleman, T. and Hinton, G. (2012).
\newblock Lecture 6.5-rmsprop: Divide the gradient by a running average of its
  recent magnitude.
\newblock {\em COURSERA: Neural networks for machine learning}, 4(2):26--31.

\bibitem[Vinyals et~al., 2017]{vinyals2017starcraft}
Vinyals, O., Ewalds, T., Bartunov, S., Georgiev, P., Vezhnevets, A.~S., Yeo,
  M., Makhzani, A., K{\"u}ttler, H., Agapiou, J., Schrittwieser, J., et~al.
  (2017).
\newblock Starcraft ii: a new challenge for reinforcement learning.
\newblock {\em arXiv preprint arXiv:1708.04782}.

\bibitem[Von~Neumann and Burks, 1966]{neumann}
Von~Neumann, J. and Burks, A.~W. (1966).
\newblock Theory of self-reproducing automata.
\newblock page~8. Urbana: University of Illinois Press.

\bibitem[Wang et~al., 2011]{wang2011self}
Wang, T., Sha, R., Dreyfus, R., Leunissen, M.~E., Maass, C., Pine, D.~J.,
  Chaikin, P.~M., and Seeman, N.~C. (2011).
\newblock Self-replication of information-bearing nanoscale patterns.
\newblock {\em Nature}, 478(7368):225.

\bibitem[Zykov et~al., 2005]{lipson}
Zykov, V., Mytilinaios, E., Adams, B., and Lipson, H. (2005).
\newblock Robotics: Self-reproducing machines.
\newblock {\em Nature}, 435:163--164.

\end{thebibliography}

\end{document}